\documentclass[letterpaper, 10 pt, conference]{ieeeconf}  

\IEEEoverridecommandlockouts                              
\usepackage{booktabs}
\usepackage{threeparttable}
\usepackage{subcaption}
\usepackage{algorithm}
\usepackage{algpseudocode}
\usepackage{lineno,hyperref}
\usepackage{amsmath,graphicx}
\usepackage{amssymb}
\usepackage{booktabs}
\usepackage{cite}
\usepackage{tablefootnote}
\usepackage{xcolor}
\usepackage{mathtools}
\usepackage{cases}
\usepackage{caption}  
\usepackage[final]{changes}
\newtheorem{theorem}{Theorem}[section]

\newtheorem{lemma}[theorem]{Lemma}
\newtheorem{corollary}[theorem]{Corollary}
\newtheorem{definition}[theorem]{Definition}
\newtheorem{assumption}[theorem]{Assumption}
\newtheorem{remark}[theorem]{Remark}

\newcommand\bE{\mathbb{E}}

\newcommand\teta{\tilde{\eta}}

\newcommand\bxi{\boldsymbol{\xi}}

\newcommand\reg{\operatorname{Regret}_d}
\newcommand\regs{\operatorname{Regret}_s}
  
\newcommand{\red}[1]{{\color{red}#1}}
\newcommand{\blue}[1]{{\color{blue}#1}}
\definecolor{purple}{rgb}{0.6, 0.2, 0.8}
\newcommand{\purple}[1]{{\color{purple}#1}}

\newcommand{\teal}[1]{{\color{teal}#1}}
\usepackage{multibib}
\definecolor{UniBlue}{RGB}{83,121,170}
\definecolor{DarkGray}{RGB}{90,90,90}
\definecolor{LightGray}{RGB}{150,150,150}
\definecolor{oldTextGreen}{RGB}{115,155,15}
\definecolor{teal}{RGB}{100, 200,10}
\definecolor{oldOcean}{RGB}{23,142,189}
\definecolor{Ocean}{RGB}{30,106,181}
\definecolor{BG}{RGB}{215,215,215}
\definecolor{darkred}{RGB}{204,41,0}

\usepackage{hyperref}
\newcommand{\zjj}[1]{{\color{darkred}#1}}
\allowdisplaybreaks
\fontsize{8}{9}\selectfont

\newcommand{\Lingzhi}[1]{{\scriptsize\textbf{\color{blue}Lingzhi: #1}}}

\title{\LARGE \bf
Differentially Private Online Federated Learning with Correlated Noise
}

\author{Jiaojiao Zhang, Linglingzhi Zhu and Mikael Johansson
\thanks{This work is supported in part by the funding from Digital Futures and VR under the contract 2019-05319.}
\thanks{J. Zhang and M. Johansson are with the Division of Decision and Control Systems, School of Electrical Engineering and Computer
Science, KTH Royal Institute of Technology, SE-100 44 Stockholm, Sweden. 
{\tt\small \{jiaoz,mikaelj\}@kth.se  }}%
\thanks{L. Zhu is with the Department of Systems
Engineering and Engineering Management, The Chinese University of Hong Kong
{\tt\small lzzhuling@gmail.com}}%
}

\begin{document}

\maketitle
\thispagestyle{empty}
\pagestyle{empty}

\begin{abstract}
We introduce a novel differentially private algorithm for online federated learning that employs temporally correlated noise to enhance utility while ensuring privacy of continuously released models. To address challenges posed by DP noise and local updates with streaming non-iid data, we develop a perturbed iterate analysis to control the impact of the DP noise on the utility. Moreover, we demonstrate how the drift errors from local updates can be effectively managed under a quasi-strong convexity condition. Subject to an $(\epsilon, \delta)$-DP budget, we establish a dynamic regret bound over the entire time horizon, quantifying the impact of key parameters and the intensity of changes in dynamic environments. Numerical experiments confirm the efficacy of the proposed algorithm.
\end{abstract}

\section{INTRODUCTION}

In this paper, we focus on online federated learning (OFL)~\cite{mitra2021online,wang2023linear,liu2023differentially}, a framework that combines the principles of federated learning (FL) and online optimization to address the challenges of real-time data processing across distributed data resources. In OFL, a central  \emph{server} coordinates multiple \emph{learners}, each interacting with streaming \emph{clients} as they arrive sequentially. The client data is used collaboratively to improve the utility of 
all learners~\cite{kairouz2021advances, qin2022decentralized}. OFL is particularly relevant for applications that require immediate decision-making. One example could be a hospital network where individual hospitals act as learners and their patients serve as clients; see Fig.~1. By enabling real-time model updates using new patient data, hospitals can offer instant health treatment or advice. Similar scenarios arise in predictive maintenance, anomaly detection, and recommendation systems.

Unlike \deleted{the} offline FL, where \replaced{the data sets that different actors have access to}{clients} are fixed before learning begins~\cite{bubeck2011introduction}, 
streaming data \replaced{that arrives at different time steps}{frequently displays characteristics indicating that data collected at different time steps} is \added{typically} not independent and identically distributed\deleted{ (non-iid)}, even on the same learner. Considering the \replaced{possibly}{probable} substantial differences among clients associated with different learners, the data across learners also exhibits non-iid characteristics, even in the same time step~\cite{mitra2021online}. 
%
\replaced{Our}{With the streaming non-iid data, our} goal is to train a model \added{using streaming non-iid data} and release it in real-time to offer \replaced{clients access to a continuously improved service}{the clients a specific service}.

A major concern in collaborative learning is the risk of privacy leakage. 
 \replaced{Clients that participate in the online learning process need assurance that their sensitive private data is not exposed to others}{Every client participating in online learning wishes for the released model not to disclose their privacy}~\cite{kairouz2021practical,denisov2022improved}.
Differential privacy (DP), which typically involves adding noise to the sensitive information to guarantee the indistinguishability of outputs~\cite{dwork2008differential, le2013differentially, cao2020differentially}, is widely recognized as a standard technique for preserving and quantifying privacy.

\replaced{Most research on differentially private federated learning focuses on an offline setting where privacy-preserving noise is added independently across iterations.}{
Plenty of works study DP algorithms with independent noise for offline learning, meaning that the noise added in different rounds of iteration is independent.} This approach allows for calculating the privacy loss per iteration and then applying composition theory to compute the total privacy loss after multiple rounds of release\replaced{. However, the injection of iid noise into the learning process also reduces the utility}{but also reduces utility}~\cite{kairouz2015composition,bassily2014private}. The online setting presents additional privacy considerations. 
\deleted{Additionally, offline privacy mechanisms are no longer directly applicable in the online setting.}
Since the model on the server is continuously released and gradients are computed at points that depend on all the previous outputs, our privacy mechanism needs to accommodate \textit{adaptive continuous} release~\cite{dwork2010differential,denisov2022improved}. This means that we have to account for \deleted{ 
Considering adaptive release implies accommodating} a more powerful adversary who can select data based on all prior outputs.
\deleted{ly observed outputs. This consideration is particularly relevant in the context of online learning with non-fixed data, enhancing the robustness of the algorithm in real applications~\cite{bubeck2011introduction}.} 
While some studies have explored differentially private OFL algorithms with independent noise~\cite{liu2023differentially, han2022differentially,odeyomi2023differentially}, none of them have addressed the challenges of adaptive continuous release.

Recently, several authors have proposed algorithms that use temporally correlated noise to improve the privacy-utility trade-off in the single-machine setting~\cite{kairouz2021practical,denisov2022improved,koloskova2023convergence,koloskova2024gradient}. These methods can be represented as a binary tree~\cite{dwork2010differential}, where the privacy analysis for the entire tree supports the adaptive setting~\cite{jain2023price}. Temporally correlated noise processes can also be constructed through matrix factorization (MF), a technique originally developed for offline settings~\cite{edmonds2020power} and recently extended to adaptive continuous release~\cite{denisov2022improved,choquette2023correlated}. The matrix factorization approach introduces new degrees of freedom that can be exploited to improve the balance between utility and privacy even further~\cite{li2015matrix,koloskova2023convergence}.

While correlated noise has proven effective in single-machine online learning, its applicability to OFL scenarios has not yet been investigated. A key distinction between OFL and single-machine online learning is the use of local updates to enhance communication efficiency~\cite{li2019convergence}. However, local updates with streaming non-iid data complicate utility analysis, particularly when combined with privacy protection using correlated noise. Moreover, even in a single-machine setting without privacy protection, establishing dynamic regret bounds without assuming convexity requires new analytical techniques due to the non-uniqueness of optimal solutions.

{\bf Contribution.} \replaced{We extend}{In this paper, we adapt} temporally correlated DP noise mechanisms, \replaced{previously studied}{primarily designed} in single-machine settings, to OFL. 
\replaced{Using a perturbed iterate technique, we analyze the combined effect of correlated}{Facing the challenges arising from} DP noise\replaced{, local updates, and streaming non-iid data.}{ and local updates with streaming non-iid data, we use a perturbed iterate analysis to control the impact of DP noise on the utility.} Specifically, we construct a virtual variable by subtracting the DP noise from the actual variable generated by our algorithm, and use it as a tool to establish a dynamic regret bound for the released global model.  
Moreover, we show how the drift error caused by local updates can be managed under a quasi-strong convexity (QSC) condition\replaced{.}{, making our analysis not limited to convex scenarios.} Subject to an $(\epsilon, \delta)$-DP budget, we establish a dynamic regret bound of $\mathcal{O}\left(\tau R^{\frac{2}{3}} \frac{(\log (1/\delta)+\epsilon)}{\epsilon^2} + \tau  C_R\right)$ over the entire time horizon, where $R$ is the number of communication rounds, $\tau$ is the number of local updates, and $C_R$ is a parameter that reflects the intensity of changes in the dynamic environment. 
\replaced{Under strong convexity, we are able to remove the dependence on $C_R$ and establish a {static} regret bound in the order of}{To validate that the error caused by $C_R$ is an inherent challenge of dynamic
regret without convexity, we use a strong convexity (SC) condition and establish a static regret bound in the order of} $\mathcal{O}\left( \tau R^{\frac{3}{4}} \frac{(\log (1/\delta)+\epsilon)}{\epsilon^2}\right)$.  
Numerical experiments confirm the efficacy of our algorithm.

\textbf{Notation.} Unless otherwise specified, all variables are \(d\)-dimensional row vectors. Accordingly, loss functions map \(d\)-dimensional row vectors to real numbers. The Frobenius norm of a matrix is denoted by \(\|\cdot\|_F\), and the \(\ell_2\)-norm of a row vector is represented by \(\|\cdot\|\). The notation \([n]\) refers to the set \(\{1, \ldots, n\}\), and \(\operatorname{P}_{\mathcal{X}^{\star}}^x\) denotes the projection of \(x\) onto the set \(\mathcal{X}^{\star}\).
We use \(\operatorname{Pr}\) to denote the probability of a random event and \(\bE\) for expectation. The notation \(\bxi\sim \mathcal{N}(0, V^2)^{R\times d}\) indicates that all entries of \(\bxi\) are independent and follow the Gaussian distribution \(\mathcal{N}(0, V^2)\).
Bold symbols represent aggregated variables: \(\mathbf{x} = [x^1; \ldots; x^R] \in \mathbb{R}^{R \times d}\) aggregates vectors \(x^r \in \mathbb{R}^{1\times d}\); \(\mathbf{G} = [g^0; \ldots; g^{R-1}] \in \mathbb{R}^{R \times d}\) aggregates vectors \(g^{r-1} \in \mathbb{R}^{1\times d}\); and \(\mathbf{x}^0 = [x^0; \ldots; x^0] \in \mathbb{R}^{R \times d}\) repeats vector \(x^0 \in \mathbb{R}^{1 \times d}\).
Matrices \(\mathbf{A}, \mathbf{B}\), and \(\mathbf{C} \in \mathbb{R}^{R \times R}\) have rows \(a^r\), \(b^r\), and \(c^r\), respectively.

\section{PROBLEM FORMULATION}\label{sec-setting}

{\bf Online federated learning}:  
\begin{figure}
\centering
\includegraphics[width=5.3cm]{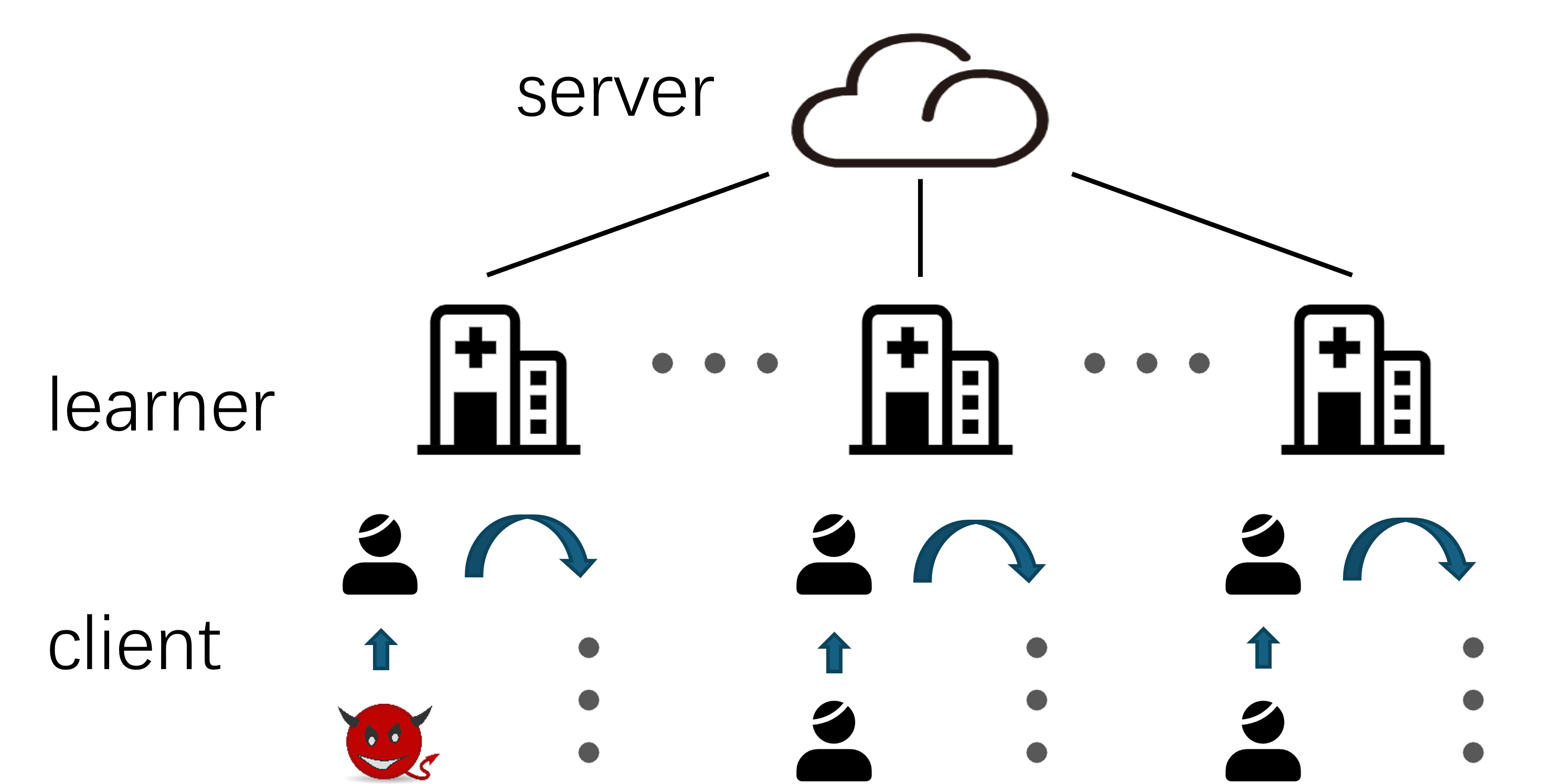}
\caption{Our OFL architecture}
\label{fig-ofl}
\end{figure}
Our OFL architecture is illustrated in Fig.~\ref{fig-ofl}. We have one server and $n$ learners, where each learner $i\in [n]$ interacts with streaming clients that arrive sequentially. We refer to the model parameters on the server and learners as the global model and local models, respectively. The learning task is for the server to coordinate all learners in training the global model online. The updated global model 
is then continuously released to the clients to provide 
instant service. 
To improve communication efficiency, learner $i$ utilizes the data of $\tau$ local clients to perform $\tau$ steps of local updates before sending the updated local model to the server. To formulate this intermittent communication, we define the entire time horizon as $\{0, 1, \ldots, \tau-1, \ldots, (R-1)\tau, \ldots, R\tau-1\}$, with communication occurring at time step $ \{r\tau: r\in[R]-1\}$, indicating a total of $R$ communication rounds spaced by $\tau$ intervals. At the $t$-th local update of the $r$-th communication round, the client, identified by $i^{r,t}\in [R\tau]$, queries the current global model $x^r$ and feeds back its data $D_i^{r,t}$ to learner $i$.  
The utility of the series of global models $\{x^r\}$ is quantified by a dynamic regret across the entire time horizon
\begin{equation*}
\reg:=\sum_{r=0}^{R-1}\sum_{t=0}^{\tau-1} \frac{1}{n}\sum_{i=1}^{n}(f_i^{r,t}(x^r)-(f^r)^{\star}). 
\end{equation*}
Here, $f_i^{r,t}(x^r)$ is the loss incurred by the global model $x^r$ on data $D_i^{r,t}$ and  
$(f^r)^{\star}$ is a dynamic optimal loss defined by
\begin{equation*}
(f^r)^{\star}:=\min_{x}  f^r(x):=\frac{1}{n\tau}\sum_{i=1}^n\sum_{t=0}^{\tau-1} f_i^{r,t} (x).
\end{equation*}
The word \textit{dynamic} means the regret metric being a difference between the loss incurred by our online algorithm and a sequence of time-varying optimal losses. In contrast, another commonly used metric is called \textit{static} regret defined by
\begin{equation*}
\regs:=\sum_{r=0}^{R-1}\sum_{t=0}^{\tau-1} \frac{1}{n}\sum_{i=1}^{n}(f_i^{r,t}(x^r)-f^r(x^{\star})), 
\end{equation*}
where $x^{\star}$ represents an optimal model, belonging to the optimal solution set $ \mathcal{X}^{\star}$  that minimizes the cumulative loss over entire data, such that
\begin{equation*}
x^{\star}\in \mathcal{X}^{\star} := \underset{x}{\operatorname{argmin}}\sum_{r=0}^{R-1}\sum_{t=0}^{\tau-1} \frac{1}{n}\sum_{i=1}^{n}f_i^{r,t}(x).  
\end{equation*} 
Unlike $\reg$, $\regs$ compares against a static optimal loss, which is made by seeing all the data in advance. 
Note that the dynamic regret is more stringent and useful than the static one in practical OFL scenarios~\cite{jiang2022distributed,eshraghi2022improving}. 

In our paper, we aim to learn a series of $\{x^r\}$ that minimizes $\reg$ while satisfying privacy constraints. 

{\bf Privacy threat model}: 
Considering the unique challenges of streaming data, we specifically allow a powerful adversary that can select data (cf. $D_i^{r,t}$) based on all previous outputs (cf. $\{x^0, \ldots, x^r\}$), highlighting the adaptive nature of our threat model. Except this, our threat model is consistent with the central DP as known in the literatures~\cite{liu2023differentially,mcmahan2017learning,geyer2017differentially}, which relies on the server to add privacy preserving noise to the aggregated updates from learners. Both the server and all learners are trustworthy, while clients are semi-honest, meaning they execute the algorithm honestly but may attempt to infer the privacy of other clients through the released global model. Additionally, we assume that the communication channels between the server and the learners are secure, preventing clients from eavesdropping.  We aim to guarantee clients' privacy under the adaptive continuous release of global models $\{x_1, \ldots, x_R\}$.

To quantify privacy leakage, we employ the concept of differential privacy. We define the aggregated dataset as $D = \{D_{i}^{r,t}: i\in[n],r\in[R]-1, t\in[\tau]-1\}$. DP is used over \textit{neighboring} datasets $D$ and $D^{\prime}$ that differ by a single entry (for instance, replacing $D_i^{r,t}$ by ${D_i^{r,t}}^{\prime}$). The formal definition of DP is provided as follows~\cite{guha2013nearly}.
\begin{definition}
A randomized mechanism $\mathcal{M}$ satisfies $(\epsilon, \delta)$-DP if for any pair of neighboring datasets $D$ and $D^{\prime}$, and for any set of outcomes $O$ within the output domain,
\begin{equation*}
\operatorname{Pr}[\mathcal{M}(D) \in {O}] \leq e^\epsilon \operatorname{Pr}\left[\mathcal{M}\left(D^{\prime}\right) \in {O}\right]+\delta.    
\end{equation*}
Here, $\mathcal{M}(D)$ represents the whole sequence of outputs generated by the mechanism $\mathcal{M}$ throughout its execution, specifically $\{x_1,\ldots, x_R\}$ in our OFL setting.  The privacy protection level is quantified by two parameters, $\epsilon$ and $\delta$, where smaller values indicate stronger privacy protection.
\end{definition}

\section{ALGORITHM}
\subsection{Proposed algorithm}
\begin{algorithm}[t]
\caption{Proposed algorithm}
\label{alg-fl}
\begin{algorithmic}[1]
\State $\textbf{Input:}$ $R$, $ \tau$, $\eta$, $\eta_g$, $\teta=\eta \eta_g \tau$, $x^{0}$, ${\bf B}$, 
$b^0=0$, and $(\epsilon,\delta)$-DP budget
\State Generate noise $\bxi\sim \mathcal{N}(0, V^2)^{R\times d}$ where $V^2$ is determined to satisfy $(\epsilon,\delta)$-DP budget
\For {$r = 0, 1, \ldots, R-1$ }
\State {\bf Learner $i$}
\State Set $z_{i}^{r,0}=x^r$
\For {$ t= 0, 1, \ldots, \tau-1$ }
\State Respond with $x^r$ to the client $i^{r,t}$
\State Obtain $D_{i}^{r,t}$ from client $i^{r,t}$
\State Update 
$z_{i}^{r,t+1}= 
z_{i}^{r,t}-\eta\nabla f_i^{r,t}(z_{i}^{r,t})$ 
\EndFor
\State Send $z_{i}^{r,\tau}$ to the server
\State {\bf Server}
\State Update  $$x^{r+1}=x^r- \teta \left(\frac{1}{\eta\tau}\left(x^r- \frac{1}{n}\sum_{i=1}^n z_{i}^{r,\tau}\right)+(b^{r+1}-b^{r})\bxi\right)$$ 
\EndFor
\State {\bf Output:} $\{x^1,\ldots, x^R\}$ 
\end{algorithmic}
\end{algorithm}

We propose a DP algorithm for OFL, as 
outlined in Algorithm~\ref{alg-fl}.  
The characteristic of our algorithm is the use of temporally correlated noise to protect privacy, alongside leveraging local updates to reduce the communication frequency between the server and learners. 

Our algorithm consists of two loops: the outer loop is indexed by $r$ for communication rounds, and the inner loop is indexed by $t$ for local updates. At the $t$-th local update of the $r$-th communication round, learner $i$ responds with the current global model $x^r$ to the client $i^{r,t}$  and obtains the client data $D_i^{r,t}$. Learner $i$ then computes the gradient of the loss $f_i^{r,t}$ at $x^r$ using data $D_i^{r,t}$ and performs one step of local update with step size $\eta$. After $\tau$ local updates, the local model $z_i^{r,\tau}$ is sent to the server. The server averages these $z_i^{r,\tau}, \forall i\in[n]$, updates the global model $x^r$ with step size $\teta$ and adds temporally correlated noise, specifically $(b^{r+1}-b^{r})\bxi$, to maintain the differential privacy of $x^{r+1}$.  

To implement our algorithm, we need to construct the matrix ${\bf B}$ and determine the variance $V^2$ of DP noise.  We will accomplish it via matrix factorization.

\subsection{Matrix factorization}
Matrix factorization, originally developed for linear counting queries~\cite{li2015matrix}, has been adapted to enhance the privacy and utility for gradient-based algorithms~\cite{kairouz2021practical}. This approach involves expressing the iterate of the gradient-based algorithm as $x^r = x^0 - \teta \sum_{\tilde{r}=0}^{r-1} g_{\tilde{r}}$, where $g_{\tilde{r}}$ is the gradient-based direction at $\tilde{r}$-th iteration, thereby determining each $x^r$ by a cumulative sum that is a specific case of linear query release~\cite[Theorem B.1]{kairouz2021practical}. Therefore, the key DP primitive is accurately estimating cumulative sums over individual gradients. 
This principle pertains to the server-side update in our OFL. Indeed, we can equivalently reformulate Line 13 of Algorithm~\ref{alg-fl} as  
\begin{equation}\label{eq-server}
\begin{aligned}
x^{r+1}=x^r-\teta \left(g^r+ (b^{r+1}-b^{r})\bxi\right), ~\forall r\in [R]-1, 
\end{aligned}
\end{equation}
where $g^r=\frac{1}{n\tau} \sum_{t=0}^{\tau-1}\sum_{i=1}^n \nabla f_i^{r,t}(z_i^{r,t})$.
Setting $b^0=0$ and repeated application of~\eqref{eq-server}  result in
\begin{equation}\label{eq-3}
{\bf x}= {\bf x}^0-\teta ({\bf A} {\bf G}+ {\bf B}\bxi ),
\end{equation}
where $\mathbf{A}$ is a lower triangular matrix with 1s on and below the diagonal. Note that we do not require ${\bf B}$ to be lower-triangular because this requirement has been relaxed by utilizing the rotational invariance of the Gaussian distribution in~\cite[Proposition 2.2]{denisov2022improved}.

Although each entry of $\bxi$ is independent, the multiplication by matrix ${\bf B}$ introduces correlations among the rows of ${\bf B}\bxi$, complicating the privacy analysis. A strategic approach involves decomposing matrix ${\bf A}$ as ${\bf A} = {\bf BC}$, thereby selecting such ${\bf B}$ to construct temporally correlated noise and then extracting ${\bf B}$ as a common factor. Substituting ${\bf {\bf A}}={\bf BC}$ into~\eqref{eq-3}, we have
\begin{equation}\label{eq-4}
{\bf x}= {\bf x}^0-\teta {\bf B} ( {\bf C} {\bf G} + \bxi).
\end{equation}
Here, the noise $\bxi$ with iid entries are added to ${\bf CG}$ and the privacy loss of~\eqref{eq-4} can be interpreted as the result of post-processing~\cite{dwork2006differential} following a single application of the Gaussian mechanism~\cite{denisov2022improved}.

The determination of the suitable matrix factorization ${\bf B}{\bf C}$ and the variance $V^2$ of DP noise relies on the analysis of utility; hence, we will address this in the next section.

\section{ANALYSIS}
Throughout this paper, we impose the following assumptions on the loss functions.    
\begin{assumption}\label{asm-smooth}
The loss function $f_i^{r,t}$ is  $L$-smooth, i.e.,
for any $x,y$, there exists a constant $L$ such that
\begin{equation*}
f_i^{r,t} (y)\leq f_i^{r,t} (x)+ \langle \nabla f_i^{r,t} (x), y-x\rangle + \frac{L}{2}\|y-x\|^2.
\end{equation*}	

\end{assumption}

\begin{assumption}\label{asm-qsc}
Consider the aggregated loss function $f^r$
and its optimal solution set $\mathcal{X}_r^{\star}:=\operatorname{argmin}_{x} f^r(x)$. The function $f^r$ is  $\mu$-quasi strongly convex, i.e.,
for any $x$, there exists a constant $\mu$ such that 
\begin{equation*}
(f^r)^\star \ge f^r(x)+ \langle \nabla f^r (x), \operatorname{P}_{\mathcal{X}_r^{\star}}^{x} -x\rangle + \frac{\mu}{2}\|\operatorname{P}_{\mathcal{X}_r^{\star}}^{x} -x\|^2.   
\end{equation*}
\end{assumption}
\begin{assumption}\label{asm-Bg}
Each loss function $f_i^{r,t}$ has bounded gradient, i.e., for any $x$, there exists a constant $B_g$ such that 
\begin{equation*}
\|\nabla f_i^{r,t}(x) \|\le B_g. 
\end{equation*}
\end{assumption}

\begin{assumption}\label{asm-regular}
For any $x^{r}$, $x^r_{\xi}$, there exists a constant $\sigma$ such that
$\|\operatorname{P}_{\mathcal{X}^{\star}}^{x^{r}}-\operatorname{P}_{\mathcal{X}^{\star}}^{x_{\xi}^{r}}\|\leq \sqrt{\sigma}\|x^r-x^r_{\xi}\|$.    
\end{assumption}

Assumption~\ref{asm-smooth} is standard in optimization literature. Assumptuon~\ref{asm-qsc} is weaker than strong convexity and a function that is QSC may be non-convex, with specific examples seen in~\cite{necoara2019linear,zhang2013gradient}. Assumption~\ref{asm-Bg} is frequently invoked in DP research to ensure bounded sensitivity~\cite{wei2023personalized,seif2020wireless}, and it aligns with the Lipschitz continuous condition of  $f_i^{r,t}$ that is common in online learning literature~\cite{denisov2022improved, mitra2021online}. Assumption~\ref{asm-regular} is a regularity condition that is necessary for our analysis since $\mathcal{X}^{\star}$ for a QSC problem may not be convex. 

With these assumptions, we are ready to establish the analysis of our algorithm on utility and privacy. All the proofs can be found in our full version~\cite{full-ofl}. 

\subsection{Utility}
Inspired by the works~\cite{koloskova2024gradient, wen2019interplay} in the single-machine setting, we employ the perturbed iterate analysis technique to control the impact of DP noise on utility. Observing the structure of temporally correlated noise $(b^{r+1}-b^r)\bxi$, which is the difference of noises at successive communication rounds, we define a virtual variable $x^{r}_{\xi}$ as
\begin{equation*}
x^{r}_{\xi}:=x^{r}+\teta b^{r}\bxi,     
\end{equation*}
and rewrite the update on the server (cf. Line 13 of Algorithm~\ref{alg-fl}) as  
\begin{equation}\label{eq-virtual}
x^{r+1}_{\xi}= x^{r}_{\xi} -\teta\cdot \frac{1}{\tau} \sum_{t=0}^{\tau-1} \frac{1}{n}\sum_{i=1}^n \nabla f_i^{r,t}(z_i^{r,t}),     
\end{equation}
where we substitute the fact that $\frac{1}{n} \sum_{i=1}^nz_i^{r,\tau}=x^r-\eta \sum_{t=0}^{\tau-1} \frac{1}{n}\sum_{i=1}^n \nabla f_i^{r,t}(z_i^{r,t})$ into Line 13 of Algorithm~\ref{alg-fl}. 
Intuitively, the virtual variable $x^{r}_{\xi}$ is designed to subtract the DP noise in $x^r$, utilizing gradient information that remains untainted by DP noise for its update, as detailed~\eqref{eq-virtual}. By analyzing the distance between $x^{r}_{\xi}$ and the optimal solution set $\mathcal{X}^{\star}$, we establish the following lemma regarding dynamic regret.  
\begin{lemma}\label{lem-u}
Under Assumptions~\ref{asm-smooth}-\ref{asm-regular}, if $\teta \le \frac{1}{8L}$, for  Algorithm~\ref{alg-fl}, we have
\begin{equation*}
\frac{\reg}{R\tau}   
\le \frac{\bE\|x_{\xi}^0-\operatorname{P}_{\mathcal{X}^{\star}}^{x_{\xi}^{0}}\|^2}{R\teta} + S^r,
\end{equation*} 
where 
\begin{equation}\notag
\begin{aligned}
&S^r: =  \left( \teta+\frac{1}{12L} + \frac{1}{\mu}\right)2 L^2\frac{\teta^2}{\eta_g^2} B_g^2
\\&+ \frac{24L(1+\sigma)\teta^2}{R} \|  {\bf B} \|_F^2dV^2+
\frac{\left(12L+{\mu}\right)C_{R}}{R}   
\end{aligned}	
\end{equation}
and $C_{R}:=\sum_{r=0}^{R-1}\bE\|\operatorname{P}_{\mathcal{X}_r^{\star}}^{x^{r}}-\operatorname{P}_{\mathcal{X}^{\star}}^{x^{r}}\|^2$.  
\end{lemma}

As shown in Lemma~\ref{lem-u}, the term  $S^r$ encapsulates several distinct sources of error including the term $B_g$ caused by the drift error from local updates, the term $V^2$ caused by DP noise, and the term $C_{R}$ caused by the dynamic environment. 

\subsection{Privacy}
Recall that ${\bf A}={\bf BC}$ and the equivalence of Algorithm~\ref{alg-fl} to ${\bf x}= {\bf x}^0-\teta {\bf B} ( {\bf C} {\bf G} + \bxi)$.  Lemma~\ref{lem-u} illuminates the impact of ${\bf B}$ and $V^2$ on utility. It is equally crucial to know the influence of ${\bf B}$, ${\bf C}$, and $V^2$ on privacy loss to determine their values. 
Based on our privacy threat model and the definition of neighboring data sets in Section~\ref{sec-setting}, we can adopt the methodology from the single-machine setting~\cite{denisov2022improved} to analyze the privacy loss incurred by the adaptive continuous release of $\{x^1,\ldots, x^R\}$. 

\begin{theorem}[Theorems 2.1 and 3.1 in~\cite{denisov2022improved}]\label{thm-dp}
Consider a lower-triangular matrix $\mathbf{A}$ of full rank and its factorization $\mathbf{A}=\mathbf{B C}$. For any neighboring sets $\mathbf{G}, \mathbf{G}' \in \mathbb{R}^{R \times d}$, it holds that $\|\mathbf{C}(\mathbf{G}-\mathbf{G}')\|_F \leq 2\gamma B_g$ where $ \gamma= \max_{r\in [R]} \|\mathbf{C}_{[:,r]} \|$ is the largest column norm of $\mathbf{C}$. Given the Gaussian noise $\bxi \sim \mathcal{N}\left(0,\frac{4\gamma^2 B_g^2(2\log (1/\delta)+\epsilon)}{\epsilon^2}\right)^{R\times d} $, if the mechanism $\mathcal{M}(\mathbf{G})=\mathbf{B}(\mathbf{C G}+\bxi)$ satisfies $(\varepsilon, \delta)$-DP in the {nonadaptive} continuous release, then $\mathcal{M}$ satisfies the same DP guarantee with the same parameters even when the rows of ${\bf G}$ are chosen adaptively.    
\end{theorem}
\begin{figure}
\centering
\includegraphics[width=5.8cm]{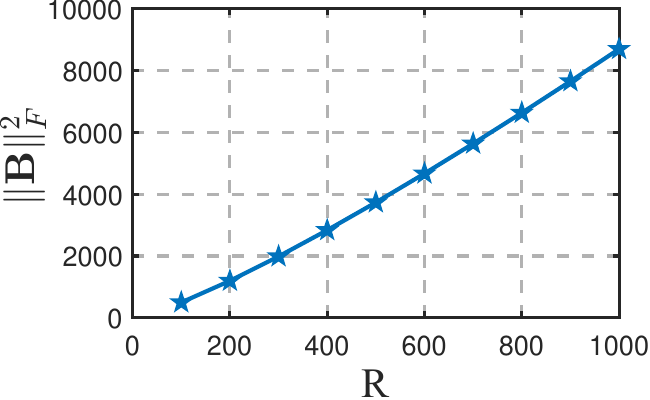}
\caption{Test to show the validity of  $\|{\bf B}\|_F^2=\mathcal{O}(R^2)$}
\label{fig-B}
\end{figure}

To achieve higher accuracy, Theorem~\ref{thm-dp} suggests a small $\gamma= \max_{r\in [R]} \|\mathbf{C}_{[:,r]} \|$ and Lemma~\ref{lem-u} suggests a small $\|{\bf B}\|^2_F$. Let us define $\mathbf{C}^{\dagger}$ as the Moore-Penrose pseudo-inverse of ${\bf C}$. Under the fact that $\mathbf{A} \mathbf{C}^{\dagger}$ yields the minimal $\ell_2$-norm solution for solving the linear equation ${\bf A}={\bf BC}$, the work~\cite{denisov2022improved} proposes to construct the matrix factors ${\bf B}$ and ${\bf C}$ through the optimization problem
\begin{equation}\label{eq-ABC}
\min _{\mathbf{C} \in \mathbf{V}, \gamma^2(\mathbf{C})=1}\|\mathbf{A} \mathbf{C}^{\dagger}\|_F^2,
\end{equation}
where $\mathbf{V}$ is a linear space of matrices. The closed-form solution of~\eqref{eq-ABC} can be found in~\cite[Theorem 3.2]{denisov2022improved}. 

By using~\eqref{eq-ABC}, we can guarantee that $\gamma=1$. For $\|{\bf B}\|_F^2$, we assume that $\|{\bf B}\|_F^2=\mathcal{O}(R^2)$ by observing that there are $R^2$ entries in ${\bf B}$, each of which are bounded. To show that $\|{\bf B}\|_F^2=\mathcal{O}(R^2)$ is reasonable (even overly conservative), we run the algorithm for solving~\eqref{eq-ABC} in~\cite[Theorem 3.2]{denisov2022improved} and plot $\|{\bf B}\|_F^2$ versus $R$ in Fig.~\ref{fig-B}, which shows that $\|{\bf B}\|_F^2$ is much smaller than $R^2$. Note that the factorization only requires prior knowledge of $R$, and it is calculated offline only once before the algorithm starts running.

\subsection{Utility-privacy}
Combining Lemma~\ref{lem-u}, Theorem~\ref{thm-dp}, and $\|{\bf B }\|_F^2=\mathcal{O}(R^2)$, we get the utility-privacy trade-off. 

\begin{theorem}\label{thm-up}
Under Assumptions~\ref{asm-smooth}-\ref{asm-regular}, if $\teta \le \frac{1}{8L}$ and $\teta= \mathcal{O}(1/R^{\frac{2}{3}})$,  Algorithm~\ref{alg-fl} subject to $(\epsilon, \delta)$-DP satisfies
\begin{equation}\notag
\begin{aligned}
\frac{\reg}{R\tau}\le\ &\mathcal{O}\left( \frac{\bE\|x_{\xi}^0-\operatorname{P}_{\mathcal{X}^{\star}}^{x_{\xi}^{0}}\|^2}{R^{\frac{1}{3}}} +  \frac{B_g^2}{\eta_g^2 R^{\frac{4}{3}}}\right.\\
& \left. + \frac{d}{R^{\frac{1}{3}}}\frac{ B_g^2(\log (1/\delta)+\epsilon)}{\epsilon^2}+\frac{C_{R}}{R}\right).   
\end{aligned}    
\end{equation}     
\end{theorem}

In Theorem~\ref{thm-up}, the errors are due to the initial error, the local updates, the DP noise, and the dynamic environment, respectively. We have the following observations. 
\begin{itemize}
    \item Given that $\teta=\eta\eta_g\tau$ and $\teta<\frac{1}{8L}$, the number of local updates $\tau$ cannot be excessively large. Such a choice would necessitate a smaller $\eta_g$, leading to a larger $\frac{B_g^2}{\eta_g^2 R^{\frac{4}{3}}}$.
    \item Our perturbed iterate analysis well-controls the impact of DP noise on utility. This is reflected in the error term caused by DP noise being $\mathcal{O}(1/R^{\frac{1}{3}})$, which decreases as $R$ increases.
    \item The term $C_{R}:=\sum_{r=0}^{R-1}\bE\|\operatorname{P}_{\mathcal{X}_r^{\star}}^{x^{r}}-\operatorname{P}_{\mathcal{X}^{\star}}^{x^{r}}\|^2$ captures that the solution set $\mathcal{X}_r^{\star}$ changes over time relative to a fixed solution set $\mathcal{X}^{\star}$,  which 
    is unavoidable for dynamic regret~\cite[Theorem 5]{zhang2017improved}.    
    Intuitively, when the environment changes very rapidly, online learning algorithms struggle to achieve high utility. 
    On the other hand, establishing a sub-linear regret bound for non-convex problems, even for static regret, presents significant challenges~\cite[Proposition 1]{jiang2022distributed}.
    Thus, to remove the dependence on $C_R$, we establish a sub-linear static regret bound under SC,  where SC simplifies the analysis by allowing us to follow steps similar to those in QSC.
\end{itemize}
  
\begin{corollary}[Static regret under SC]\label{coro}
Assume the loss function $f^r$ to be strongly convex, i.e., there exists a constant $\mu$ such that 
\begin{equation*}
\begin{aligned}
f^r(x)-f^r(x^\star)+\frac{\mu}{2}\|x-x^\star\|^2
\le \left\langle  \nabla f^r(x), x-x^\star\right\rangle, ~\forall x,
\end{aligned}      
\end{equation*}
where $x^{\star}:=\operatorname{argmin}_x \sum_{r=0}^{R-1} f^r(x)$.
Then under Assumpitions~\ref{asm-smooth},~\ref{asm-Bg} and~\ref{asm-regular}, if $\teta\le \frac{1}{10L R^{\frac{1}{2}}}$ and $\teta=\mathcal{O}(1/R^{\frac{3}{4}})$, Algorithm~\ref{alg-fl} subject to $(\epsilon, \delta)$-DP satisfies
\begin{equation*}
\begin{aligned}
\frac{\regs}{R\tau} 
\le\ & \mathcal{O}\left(\frac{\bE\|x_{\xi}^{0}-x^\star\|^2}{R^{\frac{1}{4}}} + \frac{B_g^2}{\eta_g^2 R^{\frac{3}{2}} } \right.\\
& \left. + \frac{d}{R^{\frac{1}{4}}}\frac{ B_g^2(\log (1/\delta)+\epsilon)}{\epsilon^2} + \frac{C_{R}}{R^{\frac{5}{4}}}
 \right),
\end{aligned}    
\end{equation*}  
where $C_R:=\sum_{r=0}^{R-1} \|x^{\star}-x_r^{\star}\|^2 $ and $x_r^{\star} :=\operatorname{argmin}_x f^r(x)$. 
\end{corollary}

In Corrolary~\ref{coro}, the error term caused by $C_{R}$ converges at the rate of $\mathcal{O}(1/R^{\frac{5}{4}}) $ for a static regret under SC. 

\begin{remark}
We compare our results with two closely related state-of-the-art approaches \cite{mitra2021online} and \cite{denisov2022improved}. 

The work~\cite{mitra2021online} uses local updates with non-iid streaming data for OFL but for convex problems without DP protection. The regret in~\cite{mitra2021online} is static and defined on the local models, i.e., $\regs^{\operatorname{local}}:=\sum_{r=0}^{R-1}\sum_{t=0}^{\tau-1} \frac{1}{n}\sum_{i=1}^{n}(f_i^{r,t}(z_i^{r,t})-f^r(x^{\star}))$, while ours is dynamic or static and is defined on the released global model $x^r$. 
The regret in~\cite{mitra2021online}  is upper bounded by $\mathcal{O}(\tau \log (\tau R))$ for SC case and $\mathcal{O}(\tau R^{\frac{1}{2}})$ for convex case, while our $\regs$ is bounded by $\mathcal{O}(\tau R^{\frac{3}{4}} )$ for SC case and our $\reg$ is bounded by $\mathcal{O}(\tau R^{\frac{2}{3}}+\tau C_R )$ for QSC case.  


The work~\cite{denisov2022improved} 
uses MF to guarantee DP for online learning under the adaptive continuous release but for convex problems in the single-machine setting. The regret  in~\cite[Proposition 4.1]{ denisov2022improved} is static and satisfies
\begin{equation}\label{eq-compare-dp}
\begin{aligned}
&\frac{\regs^{\operatorname{single}}}{R} \leq  
\mathcal{O}\left(\frac{E^0}{ R\eta}+\frac{\eta\|\mathbf{B}\|_F }{\sqrt{R}}  B_g V +\eta B_g^2\right)\\
& \le \mathcal{O}\left(\frac{E^0}{ R\eta}+{\eta\sqrt{R} }  B_g^2 \sqrt{ \frac{ (\log (1/\delta)+\epsilon)}{\epsilon^2}} +\eta B_g^2\right), 
\end{aligned}
\end{equation}
where $E^0$ is a constant about initial error and we substitute $\|{\bf B}\|_F=\mathcal{O}(R)$, $\gamma=1$, and $ V=\sqrt{ \frac{4 B_g^2(2\log (1/\delta)+\epsilon)}{\epsilon^2}} $ in the second inequality.
To draw a comparison with our findings, we select the step size $\eta$ in~\eqref{eq-compare-dp}. By setting $\mathcal{O}\left(\frac{1}{R\eta}\right)=\mathcal{O}(\eta \sqrt{R})$, 
$\regs^{\operatorname{single}}$ in~\cite{denisov2022improved} becomes
\begin{equation}\notag
\frac{\regs^{\operatorname{single}}}{R}\le \mathcal{O}\left(\frac{E^0}{R^{\frac{1}{4}}}+\frac{ B_g^2}{R^{\frac{1}{4}}} \sqrt{ \frac{ (\log (1/\delta)+\epsilon)}{\epsilon^2}} +\frac{B_g^2}{R^{\frac{3}{4}}} \right),
\end{equation}
which implies that $\regs^{\operatorname{single}}$ in \cite{denisov2022improved} is bounded by \(\mathcal{O}\left(R^{\frac{3}{4}}\sqrt{ \frac{ (\log (1/\delta)+\epsilon)}{\epsilon^2}} \right)\). 
Adapting our algorithm for \(n=1\), \(\tau=1\), and \(\eta_g=1\), it reduces to a single-machine algorithm. Our $\regs$ is bounded by $\mathcal{O}\left(R^{\frac{3}{4}} \frac{(\log (1/\delta)+\epsilon)}{\epsilon^2} \right)$ for SC case and our $\reg$ is bounded by $\mathcal{O}\left(R^{\frac{2}{3}} \frac{(\log (1/\delta)+\epsilon)}{\epsilon^2}+  C_R \right)$ for QSC case.  
Additionally, our proof uses QSC or SC while algorithm in~\cite{denisov2022improved} uses general convexity, which accounts for the different dependency on \(\epsilon\); ours is \((\log(1/\delta)+\epsilon)/\epsilon^2\) while that in \cite{denisov2022improved} is \(\sqrt{(\log(1/\delta)+\epsilon)/\epsilon^2}\).
\end{remark}

\section{EXPERIMENTS}
Consider a logistic regression problem 
\begin{align*}
\underset{{x}}{\operatorname{min}} \; \frac{1}{R \tau n}\sum_{r=0}^{R-1}\sum_{t=0}^{\tau-1}\sum_{i=1}^n f_i^{r,t}(x),
\end{align*}
where $f_i^{r,t}(x)= \log \left(1+\exp \left(-\left({x}\mathbf{a}_{i}^{r,t} \right) b_{i}^{r,t}\right)\right)$ is the loss function on learner $i$ and $(\mathbf{a}_{i}^{r,t}, b_{i}^{r,t}) \in \mathbb{R}^{d} \times\{-1,+1\}$ is the feature-label pair. To generate data, we use the method in~\cite{li2020federated} which allows us to control the degree of heterogeneity by two parameters $(\alpha, \beta)$.  The experiments are carried out 20 times, with the results being averaged and displayed along with error bars to indicate the standard deviation.

\subsubsection{Impact of $\tau$} 
\begin{figure}
\begin{subfigure}{0.5\textwidth}
  \centering
  \includegraphics[width=6.3cm]{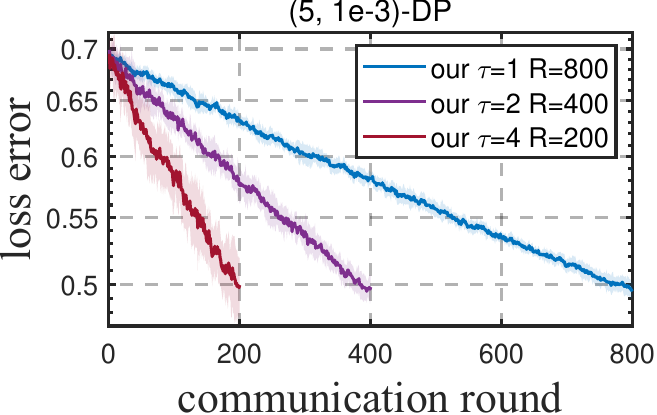}
\end{subfigure}
\caption{Impact of $\tau$}
\label{fig-tau}
\end{figure}

In the first set of experiments, we show the impact of $\tau$ on our algorithm. 
We set $n=10$ learners, with each learner receiving $800$ streaming clients. We set the parameters on data heterogeneity to $(\alpha, \beta)=(0.1, 0.1)$ and the privacy budget to $(5, 1e^{-3})$-DP. We set $\tau\in\{1, 2, 4\}$, with corresponding $R\in\{800, 400, 200\}$, so that the total data used in each test is the same. The step-sizes are the same in all tests. The results are shown in Fig.~\ref{fig-tau}. The $y$-axis represents $\operatorname{loss~error}$ of the current $x^r$ over the entire dataset,  $$\operatorname{loss~error}=\frac{1}{R \tau n}\sum_{\tilde{r}=0}^{R-1}\sum_{t=0}^{\tau-1}\sum_{i=1}^n (f_i^{\tilde{r},t}(x^r)-f_i^{\tilde{r},t}(x^{\star}))$$ 
and the $x$-axis represents the communication round.
Our algorithm achieves almost the same utility with fewer communications as $\tau$ increases while satisfying $(5,1e^{-3})$-DP.

\subsubsection{Comparison under different DP budgets}
We are unaware of any algorithm that considers the same setting as we do. The closest work appears to be the differentially private OFL algorithm in~\cite{liu2023differentially}. However, in contrast to us, their algorithm uses independent DP noise and does not consider the adaptive setting.
Its server-side update is
\begin{equation}\label{paper-batch}
x^{r+1}=x^r-\eta \left(\frac{1}{n}\sum_{i=1}^n \sum_{t=0}^{\tau-1} \nabla f_i^{r,t}(x^r) + \zeta \right), 
\end{equation}
where $\zeta\sim{\mathcal N}(0,V^2)^{1\times d}$.  
The work in~\cite{liu2023differentially} is compelling since $V^2$ required to satisfy the privacy budget can be chosen independently of $R$ and $\tau$. 
Specifically, according to \cite{liu2023differentially},  using the neighboring definition of streaming data,  the variance in~\eqref{paper-batch} is set as \(V^2={2B_g^2}/{\rho}\) with $\rho= \left(\sqrt{\epsilon+\log {1}/{\delta}}-\sqrt{\log {1}/{\delta}}\right)^2
\approx \frac{\epsilon^2}{4\log \frac{1}{\delta}} $ to ensure~\eqref{paper-batch} is $\rho$-zCDP~\cite{bun2016concentrated} and equivalently is $(\epsilon,\delta)$-DP after releasing $\{x^1,\ldots, x^R\}$. 

\begin{figure}
\begin{subfigure}{0.5\textwidth}
  \centering
  \includegraphics[width=6.3cm]{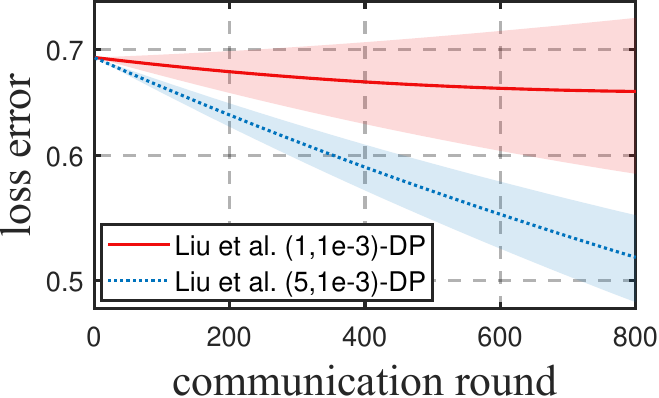}
\end{subfigure}
\vspace{1mm}
\\ 
\begin{subfigure}{0.5\textwidth}
  \centering
  \includegraphics[width=6.3cm]{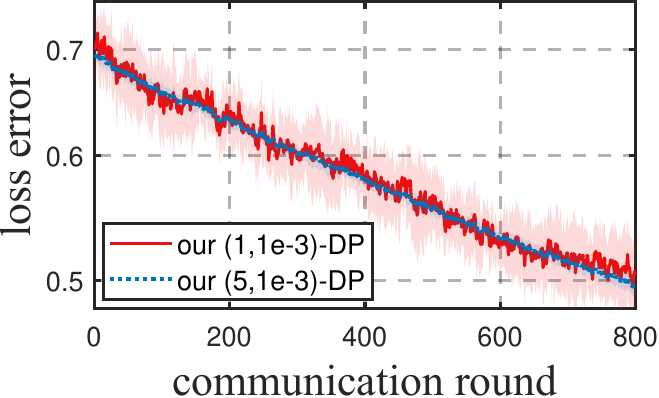}
\end{subfigure}
\caption{Comparison under different DP budgets}
\label{fig-comp}
\end{figure}

In the second set of experiments, we set $n=10$ learners, with each learner managing 8000 clients step by step, where \(\tau=10\) and \(R=800\), respectively. We set \((\alpha, \beta) = (0.1, 0.1)\).  We compare our algorithm with~\cite{liu2023differentially} under two privacy budgets \((\epsilon, \delta) \in \{(5, 1e^{-3}), (1, 1e^{-3})\}\). 

The results are shown in Fig.~\ref{fig-comp}. 
For our algorithm, we use the same step sizes under both privacy budgets, allowing our algorithm to maintain almost the same convergence rate, although the variance of the results increases when the privacy budget is stricter (i.e., $(1, 1e^{-3})$-DP). For the compared algorithm, the variance also increases as the privacy requirements become more stringent, but in addition, the algorithm has to use a smaller step-size and the convergence slows down by a lot.
%
Our algorithm performs better than~\cite{liu2023differentially}, thanks to the use of correlated noise which allows to reduce the amount of perturbation on the utility.  

\section{CONCLUSIONS}
We have proposed a DP algorithm for OFL which uses temporally correlated noise to protect client privacy under adaptive continuous release.  To overcome the challenges caused by DP noise and local updates with streaming non-iid data, we use a perturbed iterate analysis to control the impact of the DP noise on the utility. Moreover, we show how the drift error from local updates can be managed under a QSC condition. Subject to a fixed DP budget, we establish a dynamic regret bound that explicitly shows the trade-off between utility and privacy.  Numerical results demonstrate the efficiency of our algorithm.
Future research directions include extending the temporally correlated noise to scenarios where the server is not trusted,  designing optimal matrix factorization strategies tailored to improve utility, and exploring the application of the proposed algorithms to real-world scenarios in online control and signal processing.


\bibliographystyle{IEEEtran}
\bibliography{autosam}

\newpage

\appendix

\section{Fixed step size $\teta$}


The local and global updates of the proposed Algorithm~\ref{alg-fl} can be expressed as
\begin{equation}\label{eqn:illustration}
\begin{aligned}
z_i^{r,t+1} &=z_i^{r,t}-\eta{\nabla f_i^{r,t}}(z_i^{r,t}),~\forall t\in[\tau]-1, r\in[R]-1, \\
x^{r+1}&=x^r-\teta \left(\frac{1}{n\tau} \sum_{t=0}^{\tau-1}\sum_{i=1}^n \nabla f_i^{r,t}(z_i^{r,t}) + (b^{r+1}-b^{r})\bxi\right),
\end{aligned}
\end{equation}
where $z_i^{r,0}=x^r$, $\teta=\eta_g\eta\tau$, $\bxi\sim \mathcal{N}(0,V^2)^{R\times d}$. 

We aim to achieve tighter convergence by utilizing the structure of temporally correlated noise. To this end, we define 
\begin{equation*}
x^{r}_{\xi}:=x^{r}+\teta b^{r}\bxi    
\end{equation*}
and then the global updates in~\eqref{eqn:illustration} yields 
\begin{equation}\label{eq-x-xi}
x^{r+1}_{\xi}= x^{r}_{\xi} -\teta \underbrace{\frac{1}{\tau} \sum_{t=0}^{\tau-1} \frac{1}{n}\sum_{i=1}^n \nabla f_i^{r,t}(z_i^{r,t})  }_{:=g^r}.   
\end{equation}
In the following analysis, we focus on the sequence \( \{x^r_{\xi} \}_r\) instead of \( \{x^r\}_r \).

\subsection{QSC case: proof of Lemma~\ref{lem-u}}
With~\eqref{eq-x-xi}, we have
\begin{equation}\label{eq-cvx-r+1-r}
\begin{aligned}
&\bE\|x_{\xi}^{r+1}-\operatorname{P}_{\mathcal{X}^{\star}}^{x_{\xi}^{r+1}}\|^2\\
\leq\ &\bE \|x_{\xi}^r-\operatorname{P}_{\mathcal{X}^{\star}}^{x_{\xi}^{r}}-\teta g^r \|^2 \\ 
=\ & \bE\|x_{\xi}^r-\operatorname{P}_{\mathcal{X}^{\star}}^{x_{\xi}^{r}}\|^2 + {\teta^2\bE\|g^r\|^2}\\
&{-2\teta \bE\left\langle g^r,x_{\xi}^r-x^r+x^r-\operatorname{P}_{\mathcal{X}^{\star}}^{x^{r}}+\operatorname{P}_{\mathcal{X}^{\star}}^{x^{r}}-\operatorname{P}_{\mathcal{X}^{\star}}^{x_{\xi}^{r}}\right\rangle}\\
\le\ & \bE\|x_{\xi}^r-\operatorname{P}_{\mathcal{X}^{\star}}^{x_{\xi}^{r}} \|^2 \underbrace{-2\teta \bE\left\langle g^r,x^r-\operatorname{P}_{\mathcal{X}^{\star}}^{x^{r}}\right\rangle}_{\rm (IV)}+\teta^2\left(1+\frac{\theta}{\teta}\right)\underbrace{\bE\|g^r\|^2}_{\rm (V)}\\ 
&+ \frac{2\teta}{\theta}\left(\bE\|x_{\xi}^r-x^r\|^2+\bE\|\operatorname{P}_{\mathcal{X}^{\star}}^{x^{r}}-\operatorname{P}_{\mathcal{X}^{\star}}^{x_{\xi}^{r}}\|^2\right),
\end{aligned} 
\end{equation}
where the first inequality is from the optimality of $\operatorname{P}_{\mathcal{X}^{\star}}^{x_{\xi}^{r+1}}$ and the second inequality is from 
$2a^Tb\le \theta\|a\|^2+\frac{1}{\theta} \|  b \|^2$ for any $\theta>0$.
To bound the term (IV), adding and subtracting $\nabla f_i^{r,t}(x^r)$ yields
\begin{equation*}
\begin{aligned}
&{\rm (IV)}=-2\teta \bE\left\langle \frac{1}{n\tau}\sum_{t=0}^{\tau-1}\sum_{i=1}^n \nabla f_{i}^{r,t}(x^{r}),x^r-\operatorname{P}_{\mathcal{X}^{\star}}^{x^{r}}\right\rangle\\
&-2\teta \bE\left\langle \frac{1}{n\tau}\sum_{t=0}^{\tau-1}\sum_{i=1}^n (\nabla f_{i}^{r,t}(z_{i}^{r,t})-\nabla f_{i}^{r,t}(x^{r})),x^r-\operatorname{P}_{\mathcal{X}^{\star}}^{x^{r}}\right\rangle.
\end{aligned}    
\end{equation*}
We define the two terms on the right hand of the above equality as (IV.I) and (IV.II), respectively.  For (IV.I), we use the QSC to get that the cross term yields descent with the quantity
\begin{equation*}
\begin{aligned}
&{\rm (IV.I)}\\
= &  -2\teta \bE\left\langle \frac{1}{n\tau}\sum_{t=0}^{\tau-1}\sum_{i=1}^n \nabla f_{i}^{r,t}(x^{r}),x^r-\operatorname{P}_{\mathcal{X}_r^{\star}}^{x^{r}}+\operatorname{P}_{\mathcal{X}_r^{\star}}^{x^{r}}-\operatorname{P}_{\mathcal{X}^{\star}}^{x^{r}}\right\rangle\\
\leq  &-\frac{2\teta}{n\tau}\sum_{i=1}^n\sum_{t=0}^{\tau-1} \bE(f_i^{r,t}(x^r)-(f^{r})^\star)-{\teta\mu}\bE\|x^r-\operatorname{P}_{\mathcal{X}_r^{\star}}^{x^{r}}\|^2\\
& +\teta \left(\theta\bE \left\| \frac{1}{n\tau} \sum_{i=1}^n\sum_{t=0}^{\tau-1}\nabla f_i^{r,t}(x^r)
\right\|^2+\frac{1}{\theta}\bE\|\operatorname{P}_{\mathcal{X}_r^{\star}}^{x^{r}}-\operatorname{P}_{\mathcal{X}^{\star}}^{x^{r}}\|^2\right),
\end{aligned}    
\end{equation*}
where for the inequality we also use $2a^Tb\le \theta\|a\|^2+\frac{1}{\theta} \|  b \|^2$ for any $\theta>0$ and  Assumption~\ref{asm-qsc} such that
\begin{equation*}
\begin{aligned}
&\frac{1}{n\tau}\sum_{i=1}^n\sum_{t=0}^{\tau-1} (f_i^{r,t}(x)-(f^{r})^\star)+\frac{\mu}{2}\|x-\operatorname{P}_{\mathcal{X}_r^{\star}}^{x}\|^2\\
\le& \left\langle \frac{1}{n\tau}\sum_{i=1}^n\sum_{t=0}^{\tau-1} \nabla f_i^{r,t}(x), x-\operatorname{P}_{\mathcal{X}_r^{\star}}^{x}\right\rangle, ~\forall x.
\end{aligned}    
\end{equation*}
For the term (IV.II), observing that after using triangle inequality we have 
\begin{equation*}
\begin{aligned}
&{\rm (IV.II)}\\
=\ &-2\teta \bE\left\langle \frac{1}{n\tau}\sum_{t=0}^{\tau-1}\sum_{i=1}^n (\nabla f_{i}^{r,t}(z_{i}^{r,t})-\nabla f_{i}^{r,t}(x^{r})),x^r-\operatorname{P}_{\mathcal{X}^{\star}}^{x^{r}}\right\rangle \\
\le\ &  \frac{2\teta}{\mu} \bE\left\|\frac{1}{n\tau}\sum_{t=0}^{\tau-1}\sum_{i=1}^n (\nabla f_{i}^{r,t}(z_{i}^{r,t})-\nabla f_{i}^{r,t}(x^{r})) \right\|^2\notag\\
&+ \teta \mu\left(\bE\| x^r-\operatorname{P}_{\mathcal{X}_r^{\star}}^{x^{r}}\|^2+\|\operatorname{P}_{\mathcal{X}_r^{\star}}^{x^{r}}-\operatorname{P}_{\mathcal{X}^{\star}}^{x^{r}}\|^2\right).
\end{aligned}    
\end{equation*}
Substituting (IV.I) and (IV.II) into (IV), it follows that
\begin{equation*}
\begin{aligned}
{\rm (IV)}\le & 
-\frac{2\teta}{n\tau}\sum_{i=1}^n\sum_{t=0}^{\tau-1} \bE(f_i^{r,t}(x^r)-(f^{r})^\star)\\
&+\teta \theta\bE \left\| \frac{1}{n\tau} \sum_{i=1}^n\sum_{t=0}^{\tau-1}\nabla f_i^{r,t}(x^r)
\right\|^2\\
&+\teta\left(\frac{1}{\theta}+\mu\right)\bE\|\operatorname{P}_{\mathcal{X}_r^{\star}}^{x^{r}}-\operatorname{P}_{\mathcal{X}^{\star}}^{x^{r}}\|^2\\
& +\frac{2\teta}{\mu}\bE \left\|\frac{1}{n\tau}\sum_{t=0}^{\tau-1}\sum_{i=1}^n (\nabla f_{i}^{r,t}(z_{i}^{r,t})-\nabla f_{i}^{r,t}(x^{r})) \right\|^2.
\end{aligned}    
\end{equation*}
Next, for the term (V) in~\eqref{eq-cvx-r+1-r}, adding and subtracting $\nabla f_i^{r,t}(x^r)$, we have
\begin{equation*}
\begin{aligned}
&{\rm (V)}\\  
=\ & \bE\left\| \frac{1}{n\tau} \sum_{i=1}^n\sum_{t=0}^{\tau-1} (\nabla f_i^{r,t}(z_i^{r,t})-\nabla f_i^{r,t}(x^r)+\nabla f_i^{r,t}(x^r)) \right\|^2\\
\le\ & 2\bE\left\| \frac{1}{n\tau} \sum_{i=1}^n\sum_{t=0}^{\tau-1} (\nabla f_i^{r,t}(z_i^{r,t})-\nabla f_i^{r,t}(x^r))\right\|^2\\
&+2\bE \left\| \frac{1}{n\tau} \sum_{i=1}^n\sum_{t=0}^{\tau-1}\nabla f_i^{r,t}(x^r)\right\|^2.
\end{aligned}    
\end{equation*}
Substituting (IV) and (V) into~\eqref{eq-cvx-r+1-r}, we have 
\begin{equation}\label{eq-16}
\begin{aligned}
&\bE\|x_{\xi}^{r+1}-\operatorname{P}_{\mathcal{X}^{\star}}^{x_{\xi}^{r+1}}\|^2 -\bE\|x_{\xi}^{r}-\operatorname{P}_{\mathcal{X}^{\star}}^{x_{\xi}^{r}}\|^2 \\
&\le-\frac{2\teta}{n\tau}\sum_{i=1}^n\sum_{t=0}^{\tau-1} \bE(f_i^{r,t}(x^r)-(f^{r})^\star)\\
&+2c_{{\rm VI}}\cdot\underbrace{\bE\Big\| \frac{1}{n\tau} \sum_{i=1}^n\sum_{t=0}^{\tau-1} (\nabla f_i^{r,t}(z_i^{r,t})-\nabla f_i^{r,t}(x^r))\Big\|^2}_{\rm (VI)}\\
&+2c_{{\rm VII}}\cdot \underbrace{\bE \left\| \frac{1}{n\tau} \sum_{i=1}^n\sum_{t=0}^{\tau-1}\nabla f_i^{r,t}(x^r)
\right\|^2}_{\rm (VII)}\\
&+ \frac{2\teta}{\theta}\left(\bE\|x_{\xi}^r-x^r\|^2+\bE\|\operatorname{P}_{\mathcal{X}^{\star}}^{x^{r}}-\operatorname{P}_{\mathcal{X}^{\star}}^{x_{\xi}^{r}}\|^2\right)\\
&+\teta\left(\frac{1}{\theta}+\mu\right)\bE\|\operatorname{P}_{\mathcal{X}_r^{\star}}^{x^{r}}-\operatorname{P}_{\mathcal{X}^{\star}}^{x^{r}}\|^2,
\end{aligned}       
\end{equation}
where $c_{{\rm VI}}:=\teta^2\left(1+\frac{\theta}{\teta}\right) + \frac{\teta}{\mu}$ and $c_{{\rm VII}}:=\teta^2\left(1+\frac{3\theta}{2\teta}\right)$.

For the term (VI), we can transfer it to the drift error that
\begin{equation*}
{\rm (VI)}
\leq L^2\frac{1}{n\tau} \sum_{i=1}^n\sum_{t=0}^{\tau-1} \bE\|z_i^{r,t}-x^r\|^2
\end{equation*}
by repeatedly applying the local updates  
$z_i^{r,\tau-1}=z_{i}^{r,1}-\eta  \sum_{t=1}^{\tau-1} \nabla f_i^{r,t}(z_i^{r,t})$        
and $z_i^{r,0}=x^r$. 
From $\|\nabla f_i^{r,t}(x)\|\le B_g$ for any $x$, we have 
\begin{equation}\label{eq-18}
\begin{aligned}
\|z_i^{r,\tau-1}-x^r\|^2=\left\|\eta  \sum_{t=0}^{\tau-1} \nabla f_i^{r,t}(z_i^{r,t}) \right\|^2 \le \eta^2 \tau^2 B_g^2.   
\end{aligned}    
\end{equation}
Substituting~\eqref{eq-18} into (VI), we have 
\begin{equation*}
{\rm (VI)}\le L^2 \eta^2\tau^2 B_g^2.    
\end{equation*}
To handle the term (VII), we use the $L$-smoothness to transfer the gradient norm to the loss value. To this end, we use the following fact
\begin{equation*}
f_i^{r,t}(y)\le f_i^{r,t}(x)+\langle \nabla f_i^{r,t}(x), y-x \rangle+\frac{L}{2}\|y-x\|^2. 
\end{equation*}
Optimizing both hands of above inequality w.r.t. $y$, 
we get 
\begin{equation*}
\begin{aligned}
&(f^{r})^{\star}\le \min_y	\frac{1}{n\tau}\sum_{i=1}^n\sum_{t=0}^{\tau-1} f_i^{r,t}(y)\\
\leq\ & \frac{1}{n\tau}\sum_{i=1}^n\sum_{t=0}^{\tau-1}  f_i^{r,t}(x)- \frac{1}{2L}\left\|\frac{1}{n\tau}\sum_{i=1}^n\sum_{t=0}^{\tau-1} \nabla f_i^{r,t}(x) \right\|^2\\
& +\frac{L}{2}\min_{y}\left\{\left\| y-\left(x-  \frac{1}{n\tau L}\sum_{i=1}^n\sum_{t=0}^{\tau-1} \nabla f_i^{r,t}(x) \right) \right\|^2\right\}\\
=\ & \frac{1}{n\tau}\sum_{i=1}^n\sum_{t=0}^{\tau-1} f_i^{r,t}(x)- \frac{1}{2L}\left\|\frac{1}{n\tau}\sum_{i=1}^n\sum_{t=0}^{\tau-1} \nabla f_i^{r,t}(x) \right\|^2,
\end{aligned} 
\end{equation*}
which implies that
\begin{equation}\label{eq-21}
\begin{aligned}
\left\|\frac{1}{n\tau}\sum_{i=1}^n\sum_{t=0}^{\tau-1}\nabla  f_i^{r,t}(x)\right\|^2\le 2L\left(\frac{1}{n\tau}\sum_{i=1}^n\sum_{t=0}^{\tau-1} f_i^{r,t}(x)- (f^{r})^{\star} \right).
\end{aligned}    
\end{equation}
Substituting~\eqref{eq-21} into (VII), we have 
\begin{equation*}
\begin{aligned}
{\rm (VII)}
&= \bE \left\| \frac{1}{n\tau} \sum_{i=1}^n\sum_{t=0}^{\tau-1}\nabla f_i^{r,t}(x^r)\right\|^2\\
&\le \frac{1}{n\tau} \sum_{i=1}^n\sum_{t=0}^{\tau-1} 2L \bE( f_i^{r,t}(x^r) -(f^{r})^{\star}  ).
\end{aligned}    
\end{equation*}
From the derived upper bounds for (VI) and (VII) with~\eqref{eq-16}, it follows that
\begin{equation*}
\begin{aligned}
&\bE\|x_{\xi}^{r+1}-\operatorname{P}_{\mathcal{X}^{\star}}^{x_{\xi}^{r+1}}\|^2 -\bE\|x_{\xi}^{r}-\operatorname{P}_{\mathcal{X}^{\star}}^{x_{\xi}^{r}}\|^2 \\
&\le \underbrace{-\left({2\teta}- \teta^2\left(1+\frac{3\theta}{2\teta}\right)4L \right)}_{\le -\teta} \frac{1}{n\tau}\sum_{i=1}^n\sum_{t=0}^{\tau-1} \bE(f_i^{r,t}(x^r)-(f^{r})^{\star})\\
&+\left( \teta^2 \left(1+\frac{\theta}{\teta}\right) + \frac{\teta}{\mu}\right)2L^2\eta^2\tau^2 B_g^2\\
&+ \frac{2\teta}{\theta}\left(\bE\|x_{\xi}^r-x^r\|^2+\bE\|\operatorname{P}_{\mathcal{X}^{\star}}^{x^{r}}-\operatorname{P}_{\mathcal{X}^{\star}}^{x_{\xi}^{r}}\|^2\right)\\
&+\teta\left(\frac{1}{\theta}+\mu\right)\bE\|\operatorname{P}_{\mathcal{X}_r^{\star}}^{x^{r}}-\operatorname{P}_{\mathcal{X}^{\star}}^{x^{r}}\|^2.
\end{aligned}       
\end{equation*}
Choosing $\theta$ to make sure that $ -\left({2\teta}- \teta^2\left(1+\frac{3\theta}{2\teta}\right)4L \right)\le -\teta $, which can be satisfied if the following condition holds
\begin{equation}\label{eq-theta}
\teta + \frac{3}{2}\theta \le \frac{1}{4L}.    
\end{equation}

Dividing $\teta$ on both sides, we derive that 
\begin{equation}\label{eq-24}
\begin{aligned}
&\frac{1}{\teta}\bE\|x_{\xi}^{r+1}-\operatorname{P}_{\mathcal{X}^{\star}}^{x_{\xi}^{r+1}}\|^2 -\frac{1}{\teta}\bE\|x_{\xi}^{r}-\operatorname{P}_{\mathcal{X}^{\star}}^{x_{\xi}^{r}}\|^2 \\
\le & -  \frac{1}{n\tau}\sum_{i=1}^n\sum_{t=0}^{\tau-1} \bE(f_i^{r,t}(x^r)-(f^{r})^{\star})\\
&+\left( \teta \left(1+\frac{\theta}{\teta}\right) + \frac{1}{\mu}\right)2L^2\eta^2\tau^2 B_g^2\\
&+ \frac{2}{\theta}\left(\bE\|x_{\xi}^r-x^r\|^2+\bE\|\operatorname{P}_{\mathcal{X}^{\star}}^{x^{r}}-\operatorname{P}_{\mathcal{X}^{\star}}^{x_{\xi}^{r}}\|^2\right)\\
&+\left(\frac{1}{\theta}+\mu\right)\bE\|\operatorname{P}_{\mathcal{X}_r^{\star}}^{x^{r}}-\operatorname{P}_{\mathcal{X}^{\star}}^{x^{r}}\|^2.
\end{aligned}         
\end{equation}
By denoting the last three terms on the right hand of~\eqref{eq-24} as $S^r$ and reorganizing the result, we have 
\begin{equation}\label{eq-25}
\begin{aligned}
&\frac{1}{n\tau}\sum_{i=1}^n\sum_{t=0}^{\tau-1} \bE(f_i^{r,t}(x^r)-(f^{r})^{\star})\\
\le\ & \frac{1}{\teta} \bE\|x_{\xi}^{r}-\operatorname{P}_{\mathcal{X}^{\star}}^{x_{\xi}^{r}}\|^2 -  \frac{1}{\teta}\bE\|x_{\xi}^{r+1}-\operatorname{P}_{\mathcal{X}^{\star}}^{x_{\xi}^{r+1}}\|^2 + S^r.    
\end{aligned}    
\end{equation}
Repeated applying~\eqref{eq-25} and substituting Assumption~\ref{asm-regular} gives 
\begin{equation}\label{eq-26}
\begin{aligned}
& \frac{1}{R}\sum_{r=0}^{R-1}\frac{1}{n\tau}\sum_{i=1}^n\sum_{t=0}^{\tau-1} \bE(f_i^{r,t}(x^r)-(f^{r})^{\star})\\  
\le\ & \frac{\bE\|x_{\xi}^0-\operatorname{P}_{\mathcal{X}^{\star}}^{x_{\xi}^{0}}\|^2}{R\teta}  + \frac{\sum_{r=0}^{R-1} S^r}{R},
\end{aligned}    
\end{equation}
where
\begin{equation}\label{eq-27}
\begin{aligned}
&\frac{1}{R}\sum_{r=0}^{R-1} S^r
= \left( \teta \left(1+\frac{\theta}{\teta}\right) + \frac{1}{\mu}\right)2L^2\eta^2\tau^2 B_g^2\\
&+ \frac{1}{\theta R}\sum_{r=0}^{R-1}\left(2(1+\sigma)\bE\|\teta b^r\bxi\|^2+(1+\theta\mu)\bE\|\operatorname{P}_{\mathcal{X}_r^{\star}}^{x^{r}}-\operatorname{P}_{\mathcal{X}^{\star}}^{x^{r}}\|^2\right).  
\end{aligned}    
\end{equation}

Substituting~\eqref{eq-27} with the fact that 
\begin{equation*}
\bE \|b^{r}\bxi\|^2 =\bE [\|b_1^{r}\xi^1\|^2+\cdots+ \|b_r^{r}\xi^r\|^2]= \|b^r\|^2 dV^2,
\end{equation*}
where $b_r^r$ is the $r$-th entry of $b^r$, we know that~\eqref{eq-26} implies
\begin{equation*}
\begin{aligned}
& \frac{1}{R}\sum_{r=0}^{R-1}\frac{1}{n\tau}\sum_{i=1}^n\sum_{t=0}^{\tau-1} \bE(f_i^{r,t}(x^r)-(f^{r})^{\star})\\  
\le& \frac{\bE\|x_{\xi}^0-\operatorname{P}_{\mathcal{X}^{\star}}^{x_{\xi}^{0}}\|^2}{R\teta} +  \left( \teta \left(1+\frac{\theta}{\teta}\right) + \frac{1}{\mu}\right)2L^2\eta^2\tau^2 B_g^2\\
+& \frac{2(1+\sigma)\teta^2}{\theta R} \|  {\bf B} \|_F^2 dV^2+\frac{1}{\theta R}\sum_{r=0}^{R-1}(1+\theta\mu)\bE\|\operatorname{P}_{\mathcal{X}_r^{\star}}^{x^{r}}-\operatorname{P}_{\mathcal{X}^{\star}}^{x^{r}}\|^2.
\end{aligned}      
\end{equation*}
Observing the above inequality and the condition~\eqref{eq-theta}, if we choose 
\begin{equation*}
  \theta=\frac{1}{12L} \quad \text{and} \quad\teta\le \frac{1}{8L}, 
\end{equation*}
then condition~\eqref{eq-theta} hold. 
Substituting $\theta=\frac{1}{12L}$, then we have 
\begin{equation}\label{eq-utility}
\begin{aligned}
&\frac{1}{R\tau} \reg\\  
\le\ & \frac{\bE\|x_{\xi}^0-\operatorname{P}_{\mathcal{X}^{\star}}^{x_{\xi}^{0}}\|^2}{R\teta} +  \left( \teta+\frac{1}{12L} + \frac{1}{\mu}\right)2L^2\eta^2\tau^2 B_g^2\\
+& \frac{24L(1+\sigma)\teta^2}{R} \|  {\bf B} \|_F^2dV^2+
\frac{\left(12L+{\mu}\right)C_{R}}{R},
\end{aligned}    
\end{equation}
where  $C_{R}:=\sum_{r=0}^{R-1}\bE\|\operatorname{P}_{\mathcal{X}_r^{\star}}^{x^{r}}-\operatorname{P}_{\mathcal{X}^{\star}}^{x^{r}}\|^2$. Substituting $\teta=\eta \eta_g \tau$, we  complete the proof of Lemma~\ref{lem-u}. 

\subsection{QSC case: proof of Theorem~\ref{thm-up}}
In the following, we substitute the specific values given by DP analysis and choose the step sizes. Substituting the fact that $\|{\bf B}\|_F^2=\|b^1\|^2+\cdots+\|b^R\|^2\le \mathcal{O}(1+2+\cdots + R) \le \mathcal{O}(R^2)$ into~\eqref{eq-utility}, we have
\begin{equation*}
\begin{aligned}
\frac{\reg}{R\tau} 
=\mathcal{O} \left( \frac{\bE\|x_{\xi}^0-\operatorname{P}_{\mathcal{X}^{\star}}^{x_{\xi}^{0}}\|^2}{R\teta}  + \frac{\teta^2}{\eta_g^2} B_g^2 + \teta^{2} R dV^2 + \frac{C_{R}}{R}\right).       
\end{aligned}  
\end{equation*}
Next, we choose the step size $\teta=\mathcal{O}(R^{-\frac{2}{3}})$ and derive 
\begin{equation*}
\begin{aligned}
\frac{\reg}{R\tau}\le\ &\mathcal{O}\left( \frac{\bE\|x_{\xi}^0-\operatorname{P}_{\mathcal{X}^{\star}}^{x_{\xi}^{0}}\|^2}{R^{\frac{1}{3}}} +  \frac{B_g^2}{\eta_g^2 R^{\frac{4}{3}}}\right.\\
& \left. + \frac{d}{R^{\frac{1}{3}}}\frac{ B_g^2(\log (1/\delta)+\epsilon)}{\epsilon^2}+\frac{C_{R}}{R}\right),    
\end{aligned}    
\end{equation*}
where we substitute $V^2 = \frac{4\gamma^2 B_g^2(2\log (1/\delta)+\epsilon)}{\epsilon^2}$ and $\gamma=1$.  
This completes the proof of Theorem~\ref{thm-up}. 

\subsection{Strongly convex case: dynamic regret}
For the SC case, we know from~\eqref{eq-cvx-r+1-r} with the fact $x^\star=\operatorname{P}_{\mathcal{X}^{\star}}^{x^{r}}=\operatorname{P}_{\mathcal{X}^{\star}}^{x_{\xi}^{r}}$ that 
\begin{equation*}
\begin{aligned}
&\bE\|x_{\xi}^{r+1}-x^\star\|^2\\
\le\ & \bE\|x_{\xi}^r-x^\star \|^2 \underbrace{-2\teta \bE\left\langle g^r,x^r-x^\star\right\rangle}_{\rm (IV)}+\teta^2\left(1+\frac{\theta}{\teta}\right)\underbrace{\bE\|g^r\|^2}_{\rm (V)}\\ 
&+ \frac{\teta}{\theta}\bE\|x_{\xi}^r-x^r\|^2.
\end{aligned} 
\end{equation*}
The bound for (IV.I) can be simplified as follows
\begin{equation*}
{\rm (IV.I)}
\le  -\frac{2\teta}{n\tau}\sum_{i=1}^n\sum_{t=0}^{\tau-1} \bE(f_i^{r,t}(x^r)-f_i^{r,t}(x^\star)-{\teta\mu}\bE\|x^r-x^\star\|^2,
\end{equation*}
where we utilize the SC which satisfies 
\begin{equation*}
\begin{aligned}
&\frac{1}{n\tau}\sum_{i=1}^n\sum_{t=0}^{\tau-1} (f_i^{r,t}(x)-f_i^{r,t}(x^\star))+\frac{\mu}{2}\|x-x^\star\|^2\\
\le& \left\langle \frac{1}{n\tau}\sum_{i=1}^n\sum_{t=0}^{\tau-1} \nabla f_i^{r,t}(x), x-x^\star\right\rangle, ~\forall x.
\end{aligned}    
\end{equation*}
With corresponding simplified (IV.II), we derive 
\begin{equation*}
\begin{aligned}
{\rm (IV)}\le & 
-\frac{2\teta}{n\tau}\sum_{i=1}^n\sum_{t=0}^{\tau-1} \bE(f_i^{r,t}(x^r)-f_i^{r,t}(x^\star))\\
& +\frac{\teta}{\mu}\bE \left\|\frac{1}{n\tau}\sum_{t=0}^{\tau-1}\sum_{i=1}^n (\nabla f_{i}^{r,t}(z_{i}^{r,t})-\nabla f_{i}^{r,t}(x^{r})) \right\|^2.
\end{aligned}    
\end{equation*}
Following the similar analysis for the QSC case,~\eqref{eq-16} becomes 
\begin{equation}\label{eq-16-sc}
\begin{aligned}
&\bE\|x_{\xi}^{r+1}-x^{\star}\|^2 -\bE\|x_{\xi}^{r}-x^{\star}\|^2 \\
&\le-\frac{2\teta}{n\tau}\sum_{i=1}^n\sum_{t=0}^{\tau-1} \bE(f_i^{r,t}(x^r)-f_i^{r,t}(x^\star) -(f^r)^{\star} + (f^r)^{\star})\\
&+2c_{{\rm VI} }\cdot\underbrace{\bE\Big\| \frac{1}{n\tau} \sum_{i=1}^n\sum_{t=0}^{\tau-1} (\nabla f_i^{r,t}(z_i^{r,t})-\nabla f_i^{r,t}(x^r))\Big\|^2}_{{\rm (VI)} \le L^2\eta^2\tau^2B_g^2 } \\
&+2c_{{\rm VII}}\cdot \underbrace{\bE \left\| \frac{1}{n\tau} \sum_{i=1}^n\sum_{t=0}^{\tau-1}\nabla f_i^{r,t}(x^r)
\right\|^2}_{{\rm (VII)}\le\frac{1}{n\tau} \sum_{i=1}^n\sum_{t=0}^{\tau-1} 2L \bE( f_i^{r,t}(x^r) -(f^{r})^{\star}  )}\\
&+ \frac{\teta}{\theta}\bE\|x_{\xi}^r-x^r\|^2.
\end{aligned}       
\end{equation}
If condition~\eqref{eq-theta} hold, following the analysis for the above QSC case, we derive that 
\begin{equation}\label{eq-sc-dynm}
\begin{aligned}
&\frac{1}{\teta}\bE\|x_{\xi}^{r+1}-x^\star\|^2 -\frac{1}{\teta}\bE\|x_{\xi}^{r}-x^\star\|^2 \\
\le\ &-  \frac{1}{n\tau}\sum_{i=1}^n\sum_{t=0}^{\tau-1} \bE(f_i^{r,t}(x^r)-(f^r)^{\star})\\
&+\left( \teta \left(1+\frac{\theta}{\teta}\right) + \frac{1}{2\mu}\right)2L^2\eta^2\tau^2 B_g^2
+ \frac{1}{\theta}\bE\|\teta b^r \bxi\|^2\\
&+\underbrace{2 (f^r(x^{\star}) -(f^r)^{\star}  )}_{\le {L}\|x^{\star} -x_r^{\star}\|^2 }.
\end{aligned}         
\end{equation}
Set $\theta=\frac{1}{12L}$ and $\teta\le \frac{1}{8L}$ so that condition~\eqref{eq-theta} hold. Then substituting $\teta=\mathcal{O}(R^{-\frac{2}{3}})$ and $C_R:= \sum_{r=0}^{R-1}\bE\|x^{\star}-x_r^{\star}\|^2$, we have  
\begin{equation*}
\begin{aligned}
\frac{1}{R\tau}\reg 
\le\ &\mathcal{O}\left( \frac{\bE\|x_{\xi}^0-x^{\star}\|^2}{R^{\frac{1}{3}}} +  \frac{B_g^2}{\eta_g^2 R^{\frac{4}{3}}}\right.\\
& \left. + \frac{d}{R^{\frac{1}{3}}}\frac{ B_g^2(\log (1/\delta)+\epsilon)}{\epsilon^2}+\frac{C_{R}}{R}\right).    
\end{aligned}    
\end{equation*}

\subsection{Static regret under strongly convex case: proof of Corollary~\ref{coro} }
With static regret, if condition~\eqref{eq-theta} hold, \eqref{eq-16-sc} becomes 
\begin{equation*}
\begin{aligned}
&\bE\|x_{\xi}^{r+1}-x^{\star}\|^2 -\bE\|x_{\xi}^{r}-x^{\star}\|^2 \\
&\le-\frac{2\teta}{n\tau}\sum_{i=1}^n\sum_{t=0}^{\tau-1} \bE(f_i^{r,t}(x^r)-f_i^{r,t}(x^\star))\\
&+2c_{{\rm VI} }\cdot\underbrace{\bE\Big\| \frac{1}{n\tau} \sum_{i=1}^n\sum_{t=0}^{\tau-1} (\nabla f_i^{r,t}(z_i^{r,t})-\nabla f_i^{r,t}(x^r))\Big\|^2}_{ \le L^2\eta^2\tau^2B_g^2 } \\
&+2c_{{\rm VII}}\cdot \underbrace{\bE \left\| \frac{1}{n\tau} \sum_{i=1}^n\sum_{t=0}^{\tau-1}\nabla f_i^{r,t}(x^r)
\right\|^2}_{\le\frac{1}{n\tau} \sum_{i=1}^n\sum_{t=0}^{\tau-1} 2L \bE( f_i^{r,t}(x^r) -(f^{r})^{\star} + f^r(x^{\star}) -f^r(x^{\star})  )}\\
&+ \frac{\teta}{\theta}\bE\|x_{\xi}^r-x^r\|^2.
\end{aligned}       
\end{equation*}
Then~\eqref{eq-sc-dynm} becomes 
\begin{equation*}
\begin{aligned}
&\frac{1}{\teta}\bE\|x_{\xi}^{r+1}-x^\star\|^2 -\frac{1}{\teta}\bE\|x_{\xi}^{r}-x^\star\|^2 \\
\le\ &-  \frac{1}{n\tau}\sum_{i=1}^n\sum_{t=0}^{\tau-1} \bE(f_i^{r,t}(x^r)-f^r(x^{\star}))\\
&+\left( \teta \left(1+\frac{\theta}{\teta}\right) + \frac{1}{2\mu}\right)2L^2\eta^2\tau^2 B_g^2
+ \frac{1}{\theta}\bE\|\teta b^r \bxi\|^2\\
&+ \frac{2c_{{\rm VII}}}{\teta}\cdot \frac{1}{n\tau} \sum_{i=1}^n\sum_{t=0}^{\tau-1} 2L \underbrace{\bE( f_i^{r,t}(x^{\star}) -(f^{r})^{\star}  )}_{\le {L}\|x^{\star} -x_r^{\star}\|^2 }
\end{aligned}         
\end{equation*}
and consequently we obtain that
\begin{equation*}
\begin{aligned}
&\frac{1}{n\tau}\sum_{i=1}^n\sum_{t=0}^{\tau-1} \bE(f_i^{r,t}(x^r)-f^r(x^{\star}))\\
\le \ &  \frac{1}{\teta}\bE\|x_{\xi}^{r+1}-x^\star\|^2 -\frac{1}{\teta}\bE\|x_{\xi}^{r}-x^\star\|^2 \\
\ &+\mathcal{O}\left( \frac{\teta^2}{\eta_g }B_g^2 + \frac{\teta^2}{\theta}\underbrace{\|b^r\bxi\|^2}_{\le R dV^2}+ \theta \|x^{\star}-x_r^{\star}\|^2
 \right).
\end{aligned}    
\end{equation*}
Observing the above inequality and condition~\eqref{eq-theta}, if we choose 
\begin{equation*}
\theta=\teta\sqrt{R}\quad\text{and}\quad\teta\le \frac{1}{10L R^{\frac{1}{2}}},
\end{equation*}
then condition~\eqref{eq-theta} hold. Substituting $\theta=\teta \sqrt{R}$, we have
\begin{equation*}
\begin{aligned}
&\frac{1}{R} \sum_{r=0}^{R-1} \frac{1}{n\tau}\sum_{i=1}^n\sum_{t=0}^{\tau-1} \bE(f_i^{r,t}(x^r)-f^r(x^{\star}))\\
\le \ &\mathcal{O}\left(\frac{\bE\|x_{\xi}^{0}-x^\star\|^2}{R\teta} + \frac{\teta^2}{\eta_g }B_g^2 + \teta \sqrt{R} d V^2+ \teta \sqrt{R} \|x^{\star}-x_r^{\star}\|^2
 \right).
\end{aligned}    
\end{equation*}
Let $\teta=\mathcal{O}(R^{-3/4})$ and then it follows that 
\begin{equation*}
\begin{aligned}
&\frac{1}{R} \sum_{r=0}^{R-1} \frac{1}{n\tau}\sum_{i=1}^n\sum_{t=0}^{\tau-1} \bE(f_i^{r,t}(x^r)-f^r(x^{\star}))\\
\le \ &\mathcal{O}\left(\frac{\bE\|x_{\xi}^{0}-x^\star\|^2}{R^{\frac{1}{4}}} + \frac{B_g^2}{\eta_g R^{\frac{3}{2}} } + \frac{d V^2}{R^{\frac{1}{4}}} + \frac{C_{R}}{R^{\frac{5}{4}}}
 \right).
\end{aligned}    
\end{equation*}
Substituting $V^2 = \frac{4\gamma^2 B_g^2(2\log (1/\delta)+\epsilon)}{\epsilon^2}$ and $\gamma=1$, we complete the proof of Corollary~\ref{coro}.

\end{document}